\documentclass[journal]{IEEEtran}
\usepackage{graphicx}
\usepackage{amsmath}
\usepackage{amssymb}
\usepackage{booktabs}
\usepackage[sort, numbers]{natbib}
\usepackage{amssymb}
\usepackage[utf8]{inputenc} 
\usepackage[T1]{fontenc}    
\usepackage{hyperref}       
\usepackage{url}            
\usepackage{booktabs}       
\usepackage{amsfonts}       
\usepackage{nicefrac}       
\usepackage{microtype}      
\usepackage[table]{xcolor}
\usepackage{tikz}
\usepackage{xcolor}         
\usepackage{tabularx}
\usepackage{makecell}
\usepackage{utfsym}

\usepackage{bm}
\usepackage{url}            
\usepackage{amsfonts,amsmath}       
\usepackage{nicefrac}       
\usepackage{microtype}      
\usepackage{xcolor}
\usepackage{tabu}
\usepackage{multirow}       
\usepackage{graphicx}
\usepackage{graphics}
\usepackage{tabularx}
\usepackage{makecell}
\usepackage{caption}
\usepackage{wrapfig}
\usepackage{enumitem}
\usepackage{subcaption}
\usepackage{algorithm}
\usepackage{arydshln}
\usepackage{algpseudocode}
\usepackage{color, colortbl}

\usepackage{listings}
\definecolor{Gray}{gray}{0.95}
\definecolor{orange}{rgb}{0.9,0.5,0}

\usepackage{pifont}
\usepackage[capitalize]{cleveref}
\crefname{section}{Sec.}{Secs.}
\Crefname{section}{Section}{Sections}
\Crefname{table}{Table}{Tables}
\crefname{table}{Tab.}{Tabs.}
\begin{document}
	\title{Multi-Scale Implicit Transformer with Re-parameterize for Arbitrary-Scale Super-Resolution}
	\author{Jinchen Zhu, Mingjian Zhang, Ling Zheng, Shizhuang Weng
	\thanks{Corresponding author: Ling Zheng, Shizhuang Weng, Email: zhengling@ahu.edu.cn, weng\_1989@126.com,
	Organization: Anhui University, No. 111, Jiulong Road, Hefei Economic and Technological Development Zone,  Anhui,hefei, 230601,  China.}}
	
	\maketitle
	
	\begin{abstract}
		Recently, the methods based on implicit neural representations have shown excellent capabilities for arbitrary-scale super-resolution (ASSR). Although these methods represent the features of an image by generating latent codes, these latent codes are difficult to adapt for different magnification factors of super-resolution, which seriously affects their performance. Addressing this, we design Multi-Scale Implicit Transformer (MSIT), consisting of an Multi-scale Neural Operator (MSNO) and Multi-Scale Self-Attention (MSSA). Among them, MSNO obtains multi-scale latent codes through feature enhancement, multi-scale characteristics extraction, and multi-scale characteristics merging. MSSA further enhances the multi-scale characteristics of latent codes, resulting in better performance. Furthermore, to improve the performance of network, we propose the Re-Interaction Module (RIM) combined with the cumulative training strategy to improve the diversity of learned information for the network. We have systematically introduced multi-scale characteristics for the first time in ASSR, extensive experiments are performed to validate the effectiveness of MSIT, and our method achieves state-of-the-art performance in arbitrary super-resolution tasks.
	\end{abstract}
	
	\section{Introduction}
\label{sec:intro}
Single Image Super-Resolution (SISR) is a process dedicated to reconstructing high-resolution (HR) images from their low-resolution (LR) counterparts, holding significant influences in various domains, notably including satellite imaging, security monitoring and medical imaging \cite{srcnn,swinir,emt}. The advent of Convolutional Neural Networks (CNNs) and Transformers marks a transformative era for SISR \cite{srcnn,vdsr,rcan,transformer,ViT,swintrans}. Several CNN and Transformer-based methods, such as EDSR \cite{edsr}, RDN \cite{rdn} and SwinIR \cite{swinir} have achieved remarkable results. However, the majority of these methods are tailored for a fixed magnifications, which restricts their versatility and applicability in real-world scenarios where arbitrary magnification factors of SR handling is crucial.
\begin{figure}[t]
	\centering
	\begin{subfigure}{1\linewidth}
		\centering
		\includegraphics[width=1\linewidth]{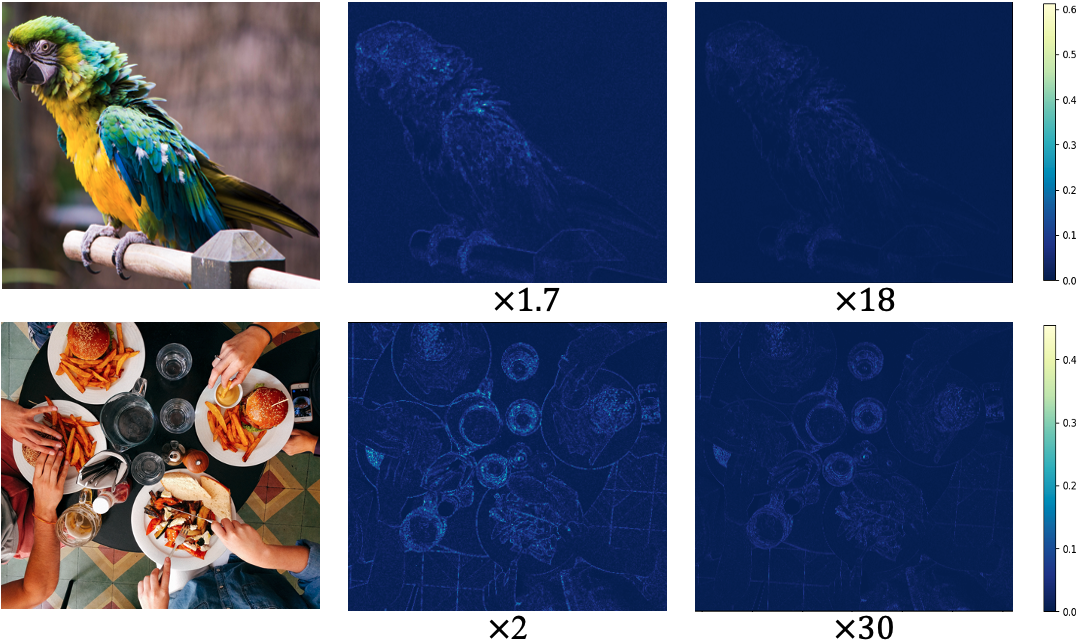}
	\end{subfigure}
	\caption{Mean error maps \cite{mem} for SR results with different magnification factors, where a brighter color indicates a larger error. For SR with large magnification factor the network has less error on texture compared to SR with small magnification factor, indicating that SR with large magnification factor focuses more on the detailed texture of the target. In contrast, at small magnification, each pixel of the image will only be slightly enlarged, thus the overall structure of the image (e.g., shape, position of the object, etc.) will be preserved, so it is more focused on the overall shape of the target.}
	\label{Msco}
\end{figure}
In recent years, there has been a remarkable increase in research interest for arbitrary scale super-resolution (ASSR) \cite{metasr,ultrasr,liif,lte}. A significant milestone was achieved by \citet{liif}, who deviated from traditional upsampling methods by employing Implicit Neural Representation (INR) \cite{nerf} to realize ASSR. The encoder first generates latent codes representing LR image features, and then the distance judgment method calculates these feature vectors in latent codes to the corresponding queried HR coordinates to produce the RGB values. Further, calculations based solely on the distance between feature vectors and HR coordinates lacks contextual information. Contemporary research \cite{clit} incorporates self-attention mechanisms in transformer \cite{transformer}  into ASSR. In addition, \citet{clit} also preliminarily found the importance of multi-scale characteristics, and therefore proposed parallel structures to take advantage of this characteristics. However, this approach introduces a huge parameter burden and is difficult to thoroughly capture multi-scale characteristics. 
Specifically, for different magnifications, the network focuses on different aspects: the network focuses more on the overall shape of the target object during SR with small magnification factors, while the network focuses more on the detailed texture of the target during large magnification, as depicted in Fig. \ref{Msco}. Therefore, it is hard for traditional latent codes to satisfy these both aspects, i.e., latent codes is difficult to adapt for different magnification factors.
Recently, convolution modulation networks have been widely used \cite{convnet,focal,conv2former}, where convolution modulation aggregates different ranges of context to obtain a stronger modeling capability than alone using convolution. Therefore, we apply the convolution modulation to ASSR to address the above problem.

Based on the above research findings, we propose an Multi-Scale Implicit Transformer (MSIT). Within MSIT framework, we propose an Multi-Scale Neural Operator (MSNO) to generate and optimize multi-scale latent codes by enriching features and applying convolution modulation. In addition, we introduce Multi-Scale Self-Attention (MSSA) to further enhance the multi-scale characteristics of latent codes.
Finally, we propose Re-Interaction Module (RIM) combined with the cumulative training strategy to learn more diverse information for the network.
The superiority and practicality of our method have been thoroughly validated through extensive experimentation in the SR domain.

	\section{Related Work}
\label{sec:relate}
\begin{figure*}[h]
	\centering
	\begin{subfigure}{1\linewidth}
		\centering
		\includegraphics[width=1\linewidth]{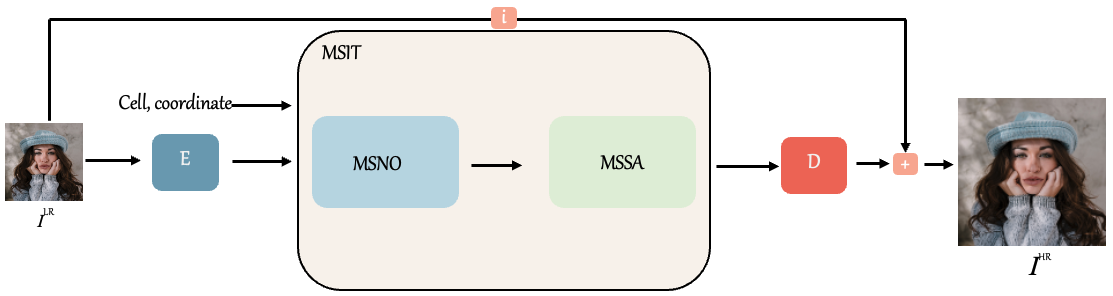}
	\end{subfigure}
	\caption{Overall architecture for continuous image SR}
	\label{MSIT}
\end{figure*}
\paragraph{Single image super-resolution}
SISR is a fundamental task in computer vision aimed at recovering HR images from LR inputs. The emergence of CNN \cite{srcnn,fsrcnn,vdsr,edsr} marks an important milestone in SISR research. Among them, SRCNN \cite{srcnn} first applied CNN to realize SISR. As the field continues to advance, a large number of complex models have emerged, such as EDSR \cite{edsr}, RDN \cite{rdn}, and RCAN \cite{rcan}. These models have demonstrated excellent capabilities in terms of SISR performance. Recently, transformer-based methods have been introduced to SISR, such as IPT \cite{ipt} and SwinIR \cite{swinir}, and they have achieved excellent results. However, these methods are limited to fixed magnifications (e.g., $\times$2, $\times$3, $\times$4), which restricts the application of SISR models in reality.

\paragraph{Arbitrary scale super-resolution based on INR}
INR is a continuous domain signal approximator rooted in multi-layer perceptrons (MLPs). It is often used for various 3D tasks, e.g., 3D rendering, 3D reconstruction \cite{3dscene_0,3dscene_1,3dscene_2,3dscene_3}. 
Recently, there has been a significant increase in the use of INRs for SR research. 
\citet{liif} first proposed the ASSR method based on local implicit neural functions in INRs, which is a new approach to produce pixel values by converting the input image into latent codes and then predicting the feature vectors in latent codes around each HR coordinate.
\citet{ultrasr} proposes a creative approach that skillfully fuses spatial coordinates and periodic coding within the framework of INRs. Subsequently, \citet{lte} introduces a Local Texture Estimator and transforms the input coordinates into the Fourier domain to this design greatly improves the expressiveness of the ASSR.
Recently, \citet{multi} and \citet{clit} preliminarily are aware of the importance of multi-scale characteristics in ASSR. \citet{multi} specifically designed multi-scale cross-fusion network as an encoder to improve the SR performance. However, this design has a fatal problem that the method is unable to use other excellent networks as encoders, e.g., EDSR \cite{edsr}, RDN \cite{rdn}, which makes the encoders difficult to capture image features efficiently. Then, \citet{clit} adopts the framework of parallelizing multiple networks to utilize multi-scale characteristics of latent codes to achieve outstanding performance. Nevertheless, the parallel structure introduces a large number of additional parameters, and this approach also is hard for latent codes to thoroughly obtain multi-scale characteristics, which makes the performance still limited. Therefore, the lack of multi-scale characteristics for latent codes has still not been effectively addressed.

\paragraph{Structural re-parameterization}
The training of deep learning models can often benefit from structural re-parameterization, which typically integrates additional branches into the architecture to boost performance \cite{resrep,scaling,acnet,repvgg}. For example, in the training phase, \citet{acnet} introduces assisted vertical and horizontal convolutional branches, which are skillfully merged back into the main branch for inference, taking advantage of enhanced representational capacity while maintaining the original model structure. Likewise, \citet{repvgg} employs identity mappings alongside $3 \times 3$ convolutions during training, later transforming these shortcuts into full $3 \times 3$ convolution branches, thereby optimizing the inference process. Nevertheless, such strategies often come with the trade-off of a heightened computational burden during the training phase. RefConv \cite{refconv} is an alternative that remaps the basic weights by $3 \times 3$ convolution to save memory usage. Considering that process of using the RefConv method is similar to the cumulative training strategy commonly used in ASSR, this led us to investigate this further. However, RefConv, relying solely on the limited interaction range of 3$\times$3 convolutions, is difficult to remap convolution kernels of different sizes effectively. 


	\begin{figure*}[t]
	\centering
	\begin{subfigure}{1\linewidth}
		\centering
		\includegraphics[width=1\linewidth]{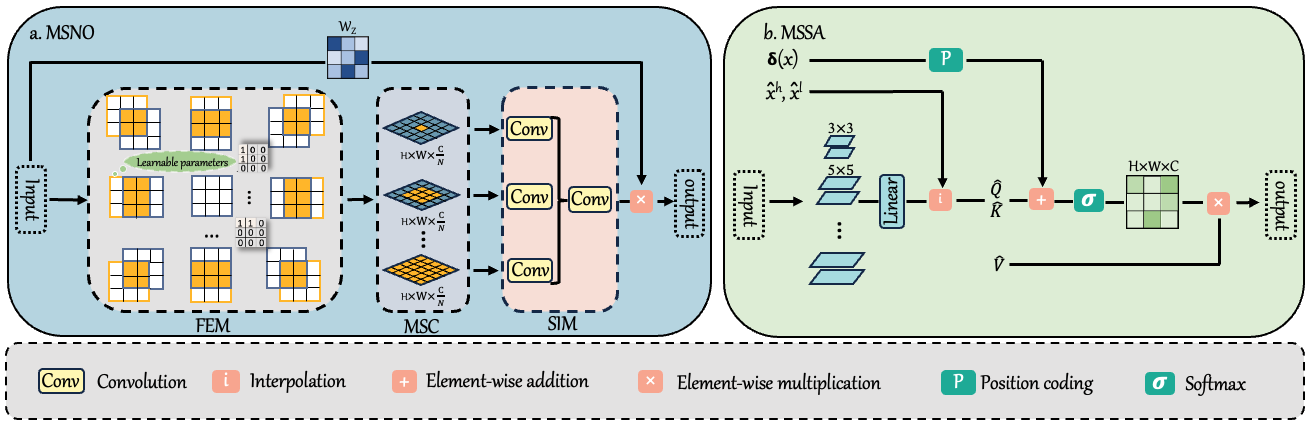}
	\end{subfigure}
	\caption{(a) Schematic Overview of MSNO Structure: Initially, the input enrichs features through FEM. Subsequently, MSC was used to obtain multi-scale latent codes. Lastly, SIM is utilized for scale mixing within the same scales and across different scales, enhancing feature diversity. (b) MSSA first obtains $Q$ and $ K$  by aggregating feature of different scales through parallel convolution. Subsequently, $Q$ and $ K$ are interpolated for $\hat\chi^h$ and $\hat \chi^l$ to obtain $\hat Q$ and $\hat K$. Subsequently, attention weights are computed by using $\hat Q$, $\hat K$, and relative coordinates. Finally, a product with the $\hat V$ is executed to generate attention latent codes $\mathcal{Z}^A$.}
	\label{MSNO}
\end{figure*}
\begin{figure*}[h]
	\centering
	\begin{subfigure}{1\linewidth}
		\centering
		\includegraphics[width=1\linewidth]{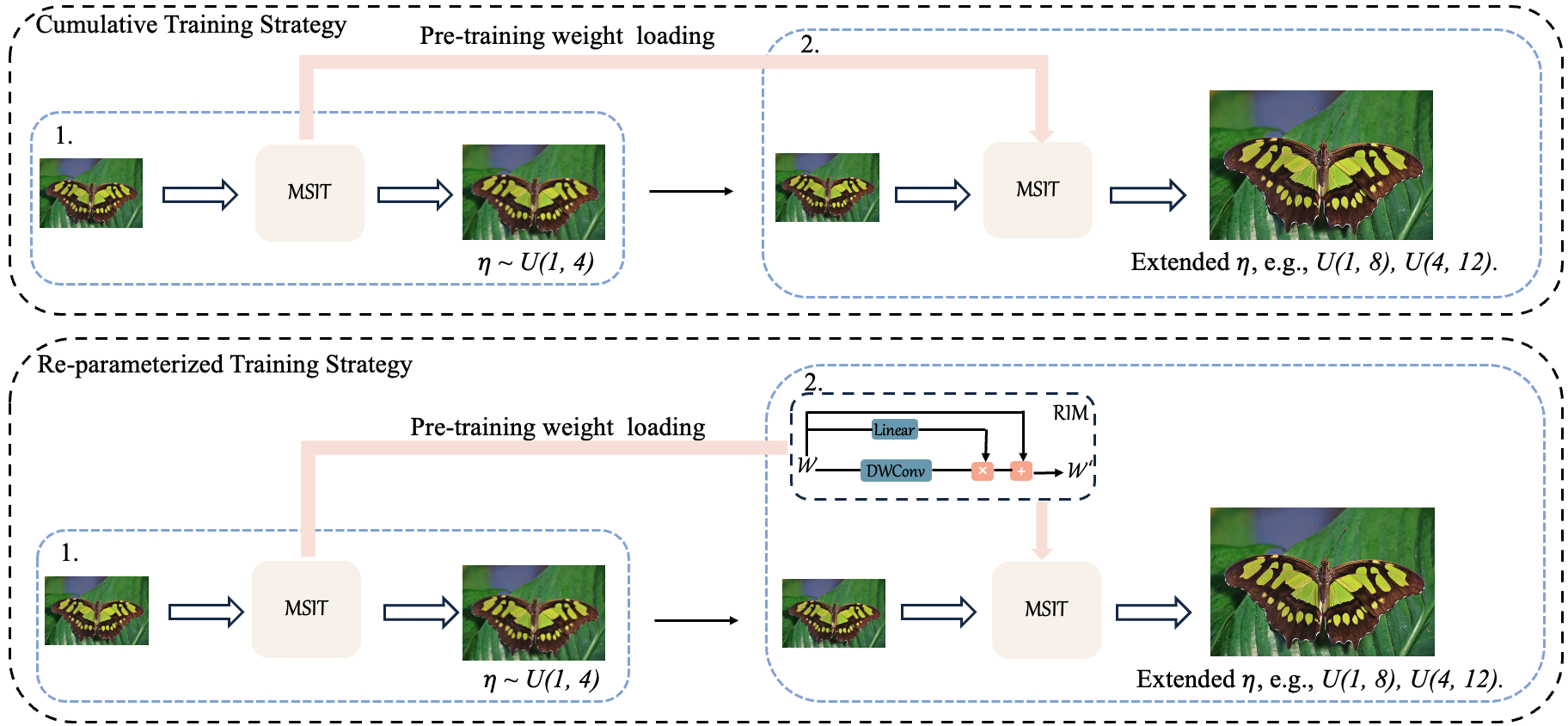}
	\end{subfigure}
	\caption{The process of re-parameterizing a training strategy, where the encoder and decoder are omitted.}
	\label{rim}
\end{figure*}
\begin{figure*}[h]
	\centering
	\begin{subfigure}{1\linewidth}
		\centering
		\includegraphics[width=1\linewidth]{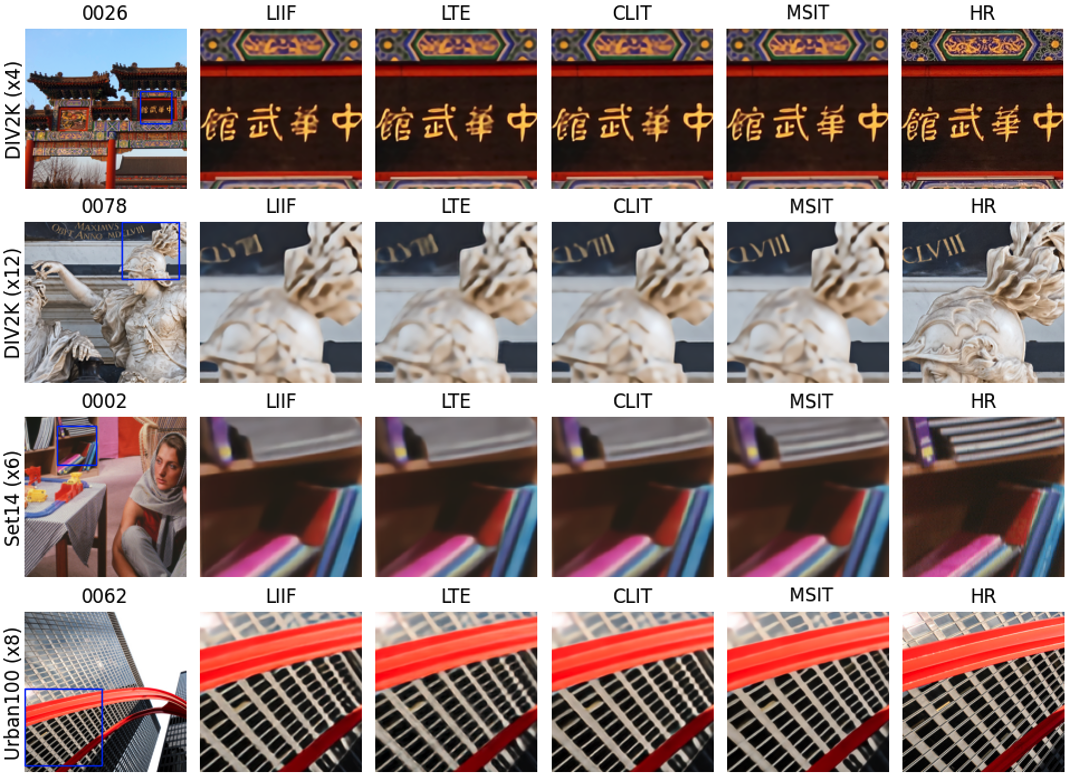}
	\end{subfigure}
	\caption{Qualitative comparison of MSIT with using RDN as the encoder.}
	\label{qualitative_0}
\end{figure*}
\begin{figure*}[t]
	\centering
	\begin{subfigure}{0.8\linewidth}
		\centering
		\includegraphics[width=1\linewidth]{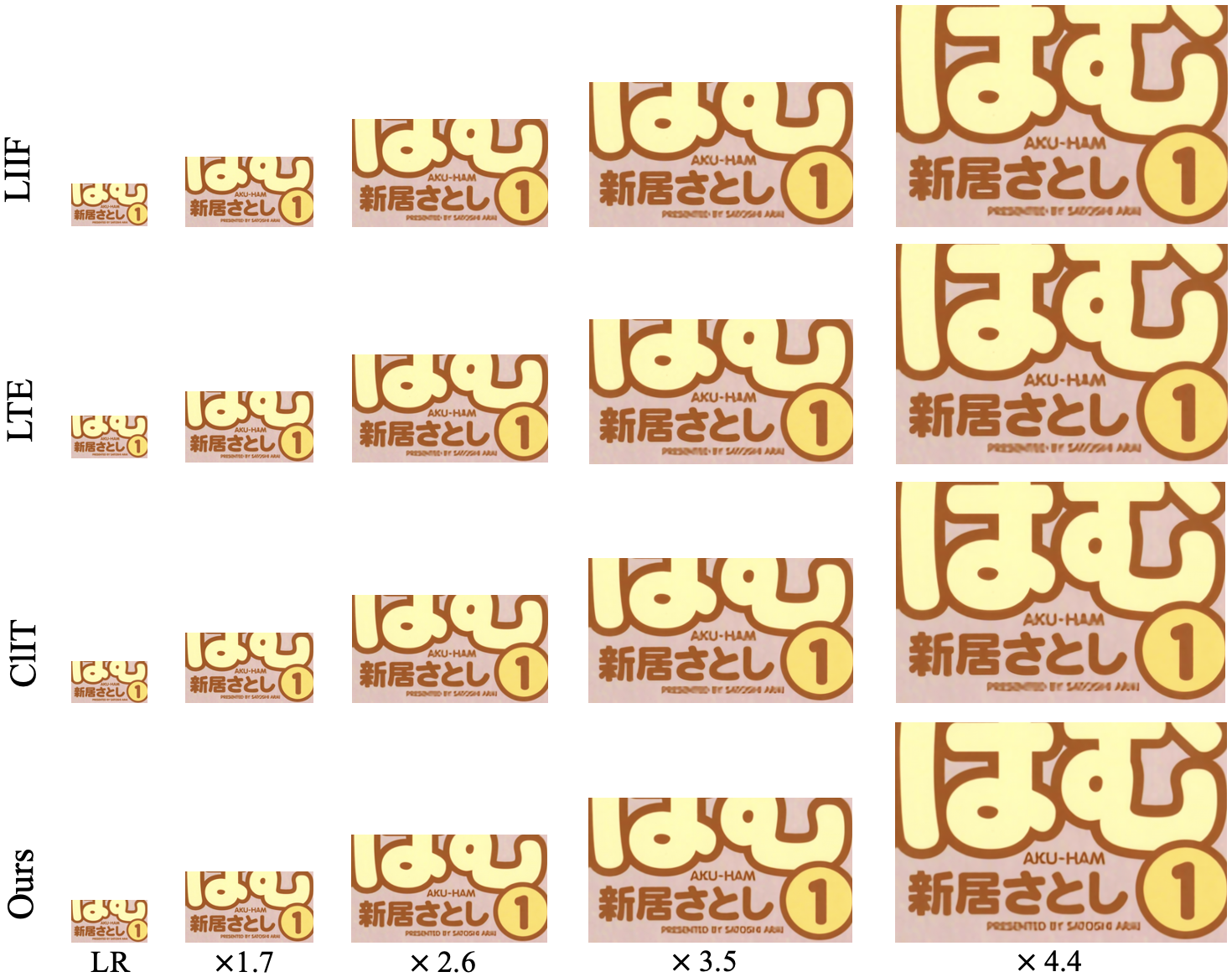}
	\end{subfigure}
	\caption{Qualitative comparison with stepwise incremental SR of MSIT with using RDN as the encoder.}
	\label{qualitative_1}
\end{figure*}
\section{Methodology}
\label{sec:meth}
In this section, we discuss the proposed ASSR network framework and the main modules.
\subsection{Overall Pipeline of Framework}
The overall framework, as shown in the Fig. \ref{MSIT}, consists of three main parts: Encoder, MSIT and Decoder. Given LR image $I_{LR}\in\mathbb{R}^{H \times W \times 3}$, LR coordinates $\chi^l :\{\chi_i^l\}_{i=1}^{i=c_g}\rightarrow\Omega_c$, and an arbitrary upsampling scale $\eta:\{\eta_h, \eta_w\}$, the HR image $I_{HR}\in\mathbb{R}^{\eta_hH \times \eta_wW \times 3}$ is recovered by generating the RGB values at HR coordinates $\chi^h:\{\chi_i^h\}_{i=1}^{i=f_g}\rightarrow\Omega_f$, where $\Omega_c, \Omega_f\subset \mathbb{R}^2 $ is the 2D coordinate space, and $f, c$ denotes the LR and HR grid sizes. The encoder $E_\psi$ first extracts the latent codes $\mathcal Z\in\mathbb{R}^{H \times W \times C}$ from $I_{lr}$, and subsequently $\mathcal Z$, $\chi^h$ and $Cell$ are fed into MSIT, where $Cell$ denotes the shape of the query pixel. In addition, we add long jump connections directly to the predictions, i.e., bilinear spatial interpolation of $I_{LR}$ and element-wise addition.
\begin{equation}
L_{HR}= \bold {D_{\psi}}\{{  \bold {MSIT}( \bold{E_{\psi}}(I_{LR}), Cell, \chi^h )}\} + I^{\uparrow}_{LR}
\end{equation}
Where spatially interpolated LR image is represented by $I^{\uparrow}_{LR}\in \mathbb{R}^{\eta_hH \times \eta_wW \times 3}$.
\subsection{Multi-Scale Implicit Transformer}
MSIT consists of two main parts, as shown in the Fig. \ref{MSNO}, MSNO and MSSA. MSNO modulates the $\mathcal Z$ generated by the encoder, and then MSSA generates the attention latent code for the HR coordinates $\chi^h$.
\paragraph{Multi-Scale Neural Operator}
In order to more systematically enable latent codes $\mathcal{Z} \in \mathbb{R}^{H \times W \times C}$ to obtain multi-scale characteristics, we introduce MSNO.
For better obtaining multi-scale latent codes, we propose the Feature Enhancement Module (FEM) in MSNO to efficiently enrich the information of latent code $\mathcal Z$ by performing spatial dimension pixel shifting operations, as shown in Fig. \ref{MSNO} (a). The input $\mathcal Z$ is first divided into eight parts along the channel dimension, where each part will be shifted along specific directions, including upward, diagonally to the upper left, diagonally to the upper right, downward, diagonally to the lower left, diagonally to the lower right, to the left, and to the right. The removed parts will be reversed to fill in the other side. In addition, FEM has a learnable shift stride size, which allows $\mathcal Z$ to enrich the feature with varying degrees. To facilitate training, FEM is implemented using convolution, where the convolution kernel is elaborately designed.
For simplicity, let's assume that the pixels are shifted to the right, with the following formula:
\begin{equation}
\resizebox{0.99\hsize}{!}{$
\mathcal Z[:, [0:W-n]:[W-n:W],:] \rightarrow \mathcal Z[:, [W-n:W]:[0:W-n],:].
$}
\end{equation}
Where $n$ denotes the learnable shift stride size. The corresponding convolution kernel is designed as follows with convolution kernel size $3\times3$:
\begin{equation}
	\begin{bmatrix}
		0 & 0 & 0 \\
		1 & 0 & 0 \\
		0 & 0 & 0 \\
	\end{bmatrix}.
\end{equation}
The purpose of these pixel shifting operations is to enrich the information of latent code with no additional parameters.

Then, we propose Multi-Scale Convolution (MSC) to efficiently obtain multi-scale latent codes. This decision comes from the fact that the convolution kernel size fundamentally affects the acquisition of spatial features \cite{convnet}. MSC is designed as a $t$ parallel convolutions, each using a specific convolution kernel, such as $ 3\times3, 5\times5, 7\times7$, and so on. Let FEM processed latent code be $\mathcal{Z'} \in \mathbb{R}^{H \times W \times C}$, and $\mathcal Z'$ is equally divided into $t$ groups:
\begin{equation}
	\mathcal Z' =  [\mathcal Z'_1, \mathcal Z'_2, ... \mathcal Z'_t].
\end{equation}
Then the MSC processing is as follows:
\begin{equation}
	\begin{split}
          MSC(\mathcal Z') &= [Conv_3(\mathcal Z'_1), Conv_5(\mathcal Z'_2), ... Conv_k(\mathcal Z'_t)]\\
                   &=[\mathcal Z''^{s_3}, \mathcal Z''^{s_5}, ... \mathcal Z''^{s_k}],
    \end{split}
\end{equation}
where $Conv_k$ denotes the $t$th parallel convolution in MSC, which $k=2t+1$ indicates the size of its processing range. $\mathcal Z''^{s_k}\in \mathbb{R}^{H \times W \times C/t}$ denotes the output of convolution with kernel size $k$.

Considering the problem of feature diversity within a single and different scales, we propose Scale Integration Module (SIM) to merge multi-scale latent codes. 
SIM sequentially mixes latent codes in single and different scales to increase feature diversity.
To capture the long range dependence in latent codes, we introduce the learnable weight matrix $\mathcal W_{{\mathcal Z}}$ to perform a Hadamard product on the output.
Since $\mathcal W_{{\mathcal Z}}$ is dynamically adjustable according to different inputs, adaptive modulation for different inputs can be realized. The process of obtaining the multi-scale latent code $\mathcal Z''$ can be summarized as follows:
\begin{equation}
	\begin{split}
		{\mathcal Z''} = \mathcal W_{ds}(\sigma(\mathcal W_{ss_3} [\mathcal Z''^{s_3}], . . . \mathcal W_{ss_k} [\mathcal Z''^{s_k}])) \times\mathcal W_{{\mathcal Z}}.\\
	\end{split}
\end{equation}
$\mathcal W_{ss}$, $\mathcal W_{ds}$ represent the operational weights used by SIM to blend feature in single and different scales, respectively, $\sigma$ represents the activation function GELU \cite{gelu}.
\paragraph{Multi-Scale Self-Attention}
MSSA is introduced to generate the attention latent code $\mathcal Z^A$ for HR coordinates. Unlike conventional SA, in MSSA, parallel convolution with different sizes of convolution kernel is used to compute multiple projection matrices. Then these projection matrices are fused together by a linear layer to further capture the multi-scale latent codes ${\mathcal Z''}$.
Specifically, MSSA takes ${\mathcal Z''}$, queried HR coordinate $\hat\chi^h:\{\hat \chi_i^h\}_{i=1}^{i=h_qw_q} \in \chi_i^h:\{\chi_i^h\}_{i=1}^{i=f_g}$ , and $\chi^l$ as inputs and generates the $\hat\chi^h$ corresponding attention latent codes ${\mathcal Z^A} \in \mathbb{R}^{h_qw_q \times C}$, where $h_q$ and $w_q$ indicate the height and width of the queried grid used to perform the calculation. The input process of MSSA can be represented by the following equation:
\begin{equation}
	\begin{split}
	{\mathcal Z^A} = Softmax(\mathcal{P}(\delta(x)) + \cfrac{\hat Q\hat K}{M} )\times \hat V,\\
\end{split}
\end{equation}
\begin{equation}
	\begin{split}
	 \mathcal{P}(\delta(\chi)) = &Linear[sin(\omega_1\delta(\chi)), cos(\omega_1\delta(\chi)), \\
&..., sin(\omega_g\delta(\chi)), cos(\omega_g\delta(\chi))],\\
	&\delta(\chi)= \hat\chi^h- \hat \chi^l.
\end{split}
\end{equation}
$\hat \chi^l :\{\hat \chi_i^l\}_{i=1}^{i=h_qw_q}$ represents the $\chi^l$ is neighborhood interpolated based on $\hat\chi^h:\{\hat \chi_i^h\}_{i=1}^{i=h_qw_q}$, and $Linear$ denotes a linear layer operation. The $\hat Q$, $\hat K$ and $\hat V$ operation process of MSSA is as follows:
\begin{equation}
	\begin{split}
		\hat Q = &[\mathcal Z'' \times \tilde{ \mathcal W_q}]_i, \\
		\hat V = &[\mathcal Z'' \times \tilde{ \mathcal W_v}]_i, \\
		&where\  \tilde{ \mathcal W_{q}} = Linear[concat(\mathcal W_{q} ^1, \mathcal W_{q} ^2, ... \mathcal W_{q} ^n)],\\
		&\ \ \ \ \ \ \ \ \ \ \ \ \tilde{ \mathcal W_{v}} = Linear[concat(\mathcal W_{v} ^1, \mathcal W_{v} ^2, ... \mathcal W_{v} ^n)].
	\end{split}
\end{equation}
Where $ \mathcal W_q^n$ and $ \mathcal W_v^n$ represents the $n$th projection matrix $q$ and $v$.
Subsequent bilinear interpolation of Q in $ \hat \chi^h$ was calculated to obtain $\hat Q$, and $V$ is neighborhood interpolated based on $\hat \chi^l$ to obtain $\hat V$. In addition, we perform $\hat \chi^l$ neighborhood interpolation computation on Q to get $
 \hat K$. We added the multi-attention mechanism and the overall computation process is as follows:
\begin{equation}
	\begin{split}
		{\mathcal Z^A} = torch.cat[Softmax(\mathcal{P}(\delta(x))_u + \cfrac{\hat Q_u\hat K_u}{M})\times \hat V_u].
	\end{split}
\end{equation}
Where $u$ represents the number of multi-head attention, and $M$ is a constant used to control the size of matrix A.
The multi-scale characteristics are effectively enhanced through MSSA to generate attention latent codes ${\mathcal{Z^A}}$, and subsequently the decoder produces RGB values based on ${\mathcal{Z^A}}$.
\subsection{Re-parameterized cumulative training strategy}
Recently, \citet{refconv} proposed RefConv to remap convolution kernels as a network re-parameterization method in advanced vision tasks, which is combined with a cumulative training-like strategy. However, this method is difficult to achieve excellent results for large convolution kernel, e.g., $7 \times 7$, $9 \times 9$, by only remapping the network weights through $3 \times 3$ depth-wise convolution (DWConv). We propose a new re-parameterization method, RIM, combined with a cumulative training strategy, as shown in Fig. \ref{rim}.
Specifically, based on the focus on the input features,  RIM further establishes connections among the weights to increase the diversity of learned information for the network. In cumulative training, RIM is used to replace the regular convolution in the network. The weights $W$ of the replaced convolution are used as input of RIM to remap a new weight $W'$, where $W$ denotes the weights obtained in the initial training stage. The inference stage will still use the replaced convolution and this new weight $W'$ will be reloaded to it. RIM mainly consists of a 3$\times$3 DWConv and linear layer. The loading weights $W$ are first remapped by DWConv. The attention map generated by the linear layer is multiplied with the output of DWConv, which allows RIM to further increase the utilization of the convolution kernel context and improve the network performance.
In addition, the network parameters are effectively reduced because the RIM has fewer parameters than the convolutions in the network. To further understand how RIM remaps the input weights, the whole process is as follows. We assume an input convolution kernel $W = [C_o, C_i, m, m]$, where $C_i$ is the input channel, $C_o$ is the output channel, and $m$ is the convolution kernel size. Firstly, the input weight $W$, reshape as [1, $C_i \times C_o$, $m$, $m$], and then perform the weight mapping, the whole process is as follows:
\begin{equation}
	\begin{split}
		W' = W + [DWConv(W) \times Linear(W)].
	\end{split}
\end{equation}
Compared to the traditional cumulative training method, replacing the regular convolution of the network with RIM not only greatly reduces the network parameters, but also improves the performance of the network for different magnification factors.

	\section{Experience}
\label{sec:expe}
\begin{table*}[t]
	\scriptsize
	\centering
	
	\resizebox{1\linewidth}{!}{
		\begin{tabu}{l|c|ccccccccc}
			\hline
			\specialrule{0em}{1pt}{0pt}
			{Method}     &{Params}      & {$\times2$}  & {$\times3$} &{$\times4$} & {$\times6$}& {$\times12$}& {$\times18$}& {$\times24$}&{$\times30$} \\
			\specialrule{0em}{1pt}{0pt}
			\hline
			\specialrule{0em}{1pt}{0pt}
			\hline
			\specialrule{0em}{1pt}{0pt}
			
			Bicubic  \cite{edsr}&- &31.01 &28.22 &26.66 &24.82 &22.27 &21.00 &20.19 &19.59                   \\
			\specialrule{0em}{1pt}{0pt}
			\hline
			\specialrule{0em}{2pt}{0pt}
			
			EDSR-baseline \cite{edsr} &1.4M&34.55 &30.90 &28.94 &- &- &- &- &-                                             \\
			EDSR-baseline-MetaSR \cite{metasr} &1.7M &34.64 &30.93 &28.92 &26.61 &23.55 &22.03 &21.06 &20.37         \\
			EDSR-baseline-LIIF \cite{liif} &1.6M&34.67 &30.96 &29.00 &26.75 &23.71 &22.17 &21.18 &20.48                          \\
			EDSR-baseline-UltraSR \cite{ultrasr} &1.6M&34.69 &31.02 &29.05 &26.81 &23.75 &22.21 &21.21 &20.51             \\
			EDSR-baseline-IPE \cite{ipe} &-&34.72 &31.01 &29.04 &26.79 &23.75 &22.21 &21.22 &20.51                         \\
			EDSR-baseline-LTE \cite{lte} &1.7M&34.72 &31.02 &29.04 &26.81 &23.78 &22.23 &21.24 &20.53                        \\
			EDSR-baseline-CLIT \cite{clit}&16.9M &\color{blue}34.82 &\color{blue}31.14 &\color{blue}29.17 &\color{blue}26.93 &\color{blue}23.85 &\color{blue}22.30 &\color{blue}21.27 &\color{blue}20.54                       \\
			EDSR-baseline-AMsIT(our) &6.0M&\color{red}{34.97} &\color{red}{31.29} &\color{red}{29.30} &\color{red}{27.03} &\color{red}{23.93} &\color{red}{22.36} &\color{red}{21.31} &\color{red}{20.59}                        \\
			
			\specialrule{0em}{1pt}{0pt}
			\hline
			\specialrule{0em}{2pt}{0pt}
			RDN \cite{rdn} &22.1M &34.94 &31.22 &29.19 &- &- &- &- &-            \\
			RDN-MetaSR \cite{metasr} &22.4M &35.00 &31.27 &29.25 &26.88 &23.73 &22.18 &21.17 &20.47             \\
			RDN-LIIF \cite{liif} &22.3M&34.99 &31.26 &29.27  &26.99 &23.89 &22.34&21.31 &20.59                            \\
			RDN-UltraSR \cite{ultrasr} &22.3M &35.00 &31.30 &29.32 &27.03 &23.73 & 22.36 &21.33 &20.61            \\
			RDN-IPE \cite{ipe}  &- &35.04 &31.32 &29.32&27.04 &23.93 &22.38 &21.34  &20.63                                  \\
			RDN-LTE \cite{lte} & 22.5M&35.04 &31.32 &29.33 &27.04 &23.95 &22.40 & 21.36 &20.64                         \\
			RDN-CLIT \cite{clit} &37.7M &\color{blue}35.10 &\color{blue}31.38 &\color{blue}29.40 &\color{blue}27.12 &\color{blue}24.01 &\color{blue}22.45 &\color{blue}21.38 &\color{blue}20.64                                                \\
			RDN-AMsIT(our) &27.9M&\color{red}{35.20} &\color{red}{31.50} &\color{red}{29.51} &\color{red}{27.21} &\color{red}{24.08} &\color{red}{22.50} &\color{red}{21.44} &\color{red}{20.68}                                             \\
			
			%

			\specialrule{0em}{1pt}{0pt}
			\hline
			
		\end{tabu}
	}
	\caption{Quantitative comparison with the SOTA method on the DIV2K validation set.
		The best results are labeled in {\color{red}red} and the second best results are labeled in {\color{blue}{blue}.}}
	\label{tab:comparison_div}
\end{table*}
\subsection{Implementation Details}
\paragraph{Datasets and Metrics}
We used the DF2K dataset to train the network \cite{div2k,edsr}. DF2K contains DIV2K and Flickr2K, where DIV2K contributes 800 high-quality images at 2K resolution, encompassing an extensive array of scenes, from serene natural landscapes to intricate urban structures, and Flickr2K added 2,650 different images capturing a variety of daily scenes from real life. This combination provides a rich dataset that is well suited for training models designed to handle various types of images. During testing, we utilize the DIV2K validation set \cite{div2k}, Set5 \cite{set5}, Set14 \cite{set14}, BSD100 \cite{BSD100} and Urban100 \cite{U100} benchmark test sets. We used the Peak Signal-to-Noise Ratio (PSNR) as the metric for evaluating our test results. Additionally, we provide the network parameters as a important indicator of network complexity for comparisons with other state-of-the-art (SOTA) methods. 

\renewcommand{\arraystretch}{1.5}
\begin{table*}[h]
	\scriptsize
	\centering
	\resizebox{1\linewidth}{!}{
		\begin{tabu}{l|ccccc|ccccc}
			\multirow{2}{*}{Method}     &\multicolumn{5}{c|}{Set5 \cite{set5}}  &\multicolumn{5}{c}{Set14 \cite{set14}}    \\ 
			&$\times2$&$\times3$&$\times4$&$\times6$&$\times8$&$\times2$&$\times3$&$\times4$&$\times6$&$\times8$ \\
			
			\hline
			
			RDN \cite{rdn} &38.24&34.71&32.47&-&-&34.01&30.57&28.81&-&-                                                                            \\
			RDN-MetaSR \cite{metasr} &38.22 &34.63 &32.38 &29.04 &26.96 &33.98 &30.54 &28.78 &26.51&24.97            \\
			RDN-LIIF \cite{liif} &38.17&34.68 &32.50 &29.15  &27.14 &33.97 &30.53 &28.80&26.64 &25.15                          \\
			RDN-UltraSR \cite{ultrasr} &38.21 &34.67 &32.49 &29.33 &27.24 &33.97 & 30.59 &28.86 &26.69 &25.25       \\
			RDN-IPE \cite{ipe}  &38.11 &34.68&32.51 &29.25&27.22 &33.94 &30.47 &28.75&26.58&25.09                      \\
			RDN-LTE \cite{lte} & 38.23&34.72&32.61 &29.32 &27.26 &34.09&30.58 &28.88 & 26.71 &25.16                           \\
			RDN-CLIT \cite{clit} &\color{blue}38.26 &\color{blue}34.80 &\color{blue}32.69 &\color{blue}29.39 &\color{blue}27.34 &\color{blue}34.21 &\color{blue}30.66 &\color{blue}28.98 &\color{blue}26.83 &\color{blue}25.35                                   \\
			RDN-MSIT(our)   &\color{red}38.32 &\color{red}34.86 &\color{red}32.74 &\color{red}29.44 &\color{red}27.36 &\color{red}34.47 &\color{red}30.89&\color{red}29.04 &\color{red}26.89 &\color{red}25.44                              \\
			
			\specialrule{0em}{1pt}{0pt}
			\cline{2-11}
			\specialrule{0em}{1pt}{0pt}
			
			\multirow{2}{*}{ }     &\multicolumn{5}{c|}{BSD100 \cite{BSD100}}  &\multicolumn{5}{c}{Urban100 \cite{U100}}    \\ 
			&$\times2$&$\times3$&$\times4$&$\times6$&$\times8$&$\times2$&$\times3$&$\times4$&$\times6$&$\times8$ \\
			
			\specialrule{0em}{1pt}{0pt}
			\cline{2-11}
			\specialrule{0em}{1pt}{0pt}
			
			RDN \cite{rdn} & 32.34&29.26&27.72&-&-&32.89& 28.80&26.61&-&-                                                                            \\
			RDN-MetaSR \cite{metasr} & 32.33 &29.26 &27.71 &25.90 &24.83  &32.92&28.82 &26.55 &23.99&22.59           \\
			RDN-LIIF \cite{liif} &32.32&29.26&27.74&25.98&24.91 &32.87&28.82&26.68&24.20&22.79                              \\
			RDN-UltraSR \cite{ultrasr} &32.35&29.29&27.77& 26.01&24.96 &32.97&28.92& 26.78& 24.30&22.87            \\
			RDN-IPE \cite{ipe}  &32.31&29.28& 27.76& 26.00&24.93&32.97&28.82&26.76&24.26&22.87                      \\
			RDN-LTE \cite{lte} &32.36 &29.30 &27.77 &26.01 &24.95 &33.04 &28.97&26.81 &24.28 &22.88                          \\
			RDN-CLIT \cite{clit} &\color{blue}32.39 &\color{blue}29.34&\color{blue} 27.82 &\color{blue}26.07 &\color{blue} 25.00 &\color{blue}33.13 &\color{blue}  29.04 &\color{blue}26.91 &\color{blue}24.43 &\color{blue}  23.03                                \\
			RDN-MSIT(our)   &\color{red}32.49 &\color{red}29.40 &\color{red}27.88 &\color{red}26.11 &\color{red}25.03 &\color{red}33.59&\color{red}29.43 &\color{red}27.26 &\color{red}24.74 &\color{red}23.27                                 \\
			
		\end{tabu}
	}

	\caption{Quantitative comparison with the SOTA method on the benchmark test sets. The best results are labeled in {\color{red}red} and the second best results are labeled in {\color{blue}{blue}.}}
	\label{tab: comparison_test}
\end{table*}

\paragraph{Training Setting}
Our training methodology follows previous research \cite{liif,ultrasr,lte,clit}. We begin by cropping HR image patches, which are sized at 48$\eta$ $\times$ 48$\eta$, where $\eta$ is randomly sampled from a uniform distribution $U(1,4)$ as the magnification factor. To generate the corresponding LR image patchs, we use bicubic interpolation in PyTorch \cite{pytorch} to downsample these HR image patches. Then, we subject these LR image patchs to data augmentation including random horizontal flips, vertical flips, and 90° rotations to enhance dataset diversity.
We randomly sample $48^2$ pixels (coordinate-RGB pairs) from each HR image patch to create our ground truth data. During training, we adopt the Adam optimizer in conjunction with the L1 loss function. 
The learning rate was initialized to $1 \times 10^{-5}$ and then incremented to $1 \times 10^{-4}$ after 50 epochs as a warm-up phase.  We employed cosine annealing scheduling to train the model for 1000 epochs , with a batch size of 32.

\subsection{Comparison with state-of-the-art methods}
\paragraph{Quantitative analysis}We adopt proposed method to conduct a comparative analysis with other SOTA that employ encoders based on EDSR and RDN architectures. The results are shown in Tab. \ref{tab:comparison_div}. CLIT preliminarily utilizes multi-scale characteristics through a parallel structure, resulting in suboptimal outcomes. However, it is worth noting that the large number of parameters in CLIT brings a great challenge to the computation. MSIT effectively utilizes multi-scale characteristics through MSNO and MSSA to produce excellent results at all magnifications.  For example, MSIT has a huge lead in the Urban100 \cite{U100} test set, far superior to other methods. MSIT adapts to different magnification factors of SR while achieving remarkable results, which makes it a competitive choice in SOTA for ASSR tasks.

\paragraph{Qualitative analysis}
We conducted a qualitative comparison of MSIT with other methods, as shown in Fig. \ref{qualitative_0}. For CLIT, due to the absence of official pre-trained weights, we trained the model by ourselves.  Based on the results in this figure, we find that other models such as LIIF exhibit obvious artifacts in the DIV2K validation set 0026 \cite{div2k} of $\times$4 SR results, especially in the "hua" font, which is alleviated in CLIT by utilizing the SA mechanism, but some artifacts still appear. Additionally, in the $\times$8 0078 SR results, there is significant over-smoothing, resulting in unrecognizable fonts. MSIT enhances the multi-scale characteristics of the network with MSNO and MSSA, effectively mitigating these issues. In the Set14 barbara.png \cite{set14} of $\times$6 SR, the other methods also showed obvious over-smoothing phenomenon, and the details are unclear enough, which seriously affects the practicability. MSIT recovered these images more clearly in the detailed texture. In the Urban100 0062 \cite{U100} of $\times$8 SR,  LIIF recovered the high-frequency edges insufficient, while LTE mitigates this problem by introducing a Fourier mechanism, but the results are still poor. MSIT achieves the best results  by extracting the texture and overall structure of the target object. The results of using stepwise incremental SR compared to other methods are shown in Fig. \ref{qualitative_1}. The input image is shown on the left, and different magnification factors of SR are used: $\times$1.7, $\times$2.6, $\times$3.5, $\times$4.4. It can be seen that MSIT effectively recovers the details of the input, such as the "HAM" in the image as well as the text at the bottom, in comparison with other methods.
\begin{table}[h]
	\small
	\setlength{\tabcolsep}{0.6mm}
	\centering
	
	\begin{tabu}{|c|c|c|c|c|cccc|}
		
		\cline{1-9}
		\multirow{2}{*}{MSNO} & \multirow{2}{*}{MSSA} & \multirow{2}{*}{RIM}& \multirow{2}{*}{Ref \cite{refconv}}  & \multirow{2}{*}{Params}& \multicolumn{4}{c|}{DIV2K val 100 \cite{div2k}}                                                    \\ 
		&                         &                                              &                       &                      & \multicolumn{1}{c}{$\times2$} & \multicolumn{1}{c}{$\times4$} & \multicolumn{1}{c}{$\times6$} & \multicolumn{1}{c|}{$\times12$} \\ \hline
		\specialrule{0em}{1pt}{0pt}
		\cline{1-9}
		\specialrule{0em}{1pt}{0pt} 
		\usym{2613}                                                    &       \usym{2613}               &  \usym{2613}       &    \usym{2613}                   & \multicolumn{1}{c|}{8.4M}  & \multicolumn{1}{c|}{34.80}  & \multicolumn{1}{c|}{29.11}  & \multicolumn{1}{c|}{26.84} & 23.75  \\ \cline{1-9} 
		\checkmark               &               \usym{2613}      &  \usym{2613}         &  \usym{2613}      &\multicolumn{1}{c|}{8.4M}   & \multicolumn{1}{c|}{34.87}  & \multicolumn{1}{c|}{29.15}  & \multicolumn{1}{c|}{26.88}  &23.79   \\ \cline{1-9} 
		\checkmark                            &    \checkmark   	& \usym{2613}    & \usym{2613}         &      \multicolumn{1}{c|}{8.1M}  & \multicolumn{1}{c|}{34.87}  & \multicolumn{1}{c|}{29.16}  & \multicolumn{1}{c|}{26.89}  & 23.81  \\ \cline{1-9} 
		\checkmark                          &     \checkmark   & \checkmark   &\usym{2613}  &   \multicolumn{1}{c|}{$\textbf{6.0M}$}  & \multicolumn{1}{c|}{$\textbf{34.91}$}  & \multicolumn{1}{c|}{$\textbf{29.19}$}  & \multicolumn{1}{c|}{$\textbf {26.91}$}  &$\textbf{23.83}$\\ \cline{1-9}
		\checkmark                          &     \checkmark    &\usym{2613}   &\checkmark   &   \multicolumn{1}{c|}{$\textbf{6.0M}$}  & \multicolumn{1}{c|}{$\textbf{34.91}$}  & \multicolumn{1}{c|}{$\textbf{29.19}$}  & \multicolumn{1}{c|}{26.89}  &23.81   \\ \cline{1-9}
		
	\end{tabu}
	\caption{The outcomes of individual module ablation experiments for MSIT on the DIV2K Validation dataset \cite{div2k} are presented in the results table, where "$\checkmark$" signifies the utilization of the respective module, and upsampling scale used by Ref and RIM is $\eta \thicksim U (1, 4)$. The best performing results are highlighted in $\textbf{bold}$.}
	\label{tab_comparison_ab1}
\end{table}
\renewcommand{\arraystretch}{1.5}
\begin{table*}[h]
	\scriptsize
	\centering
	
	\resizebox{1\linewidth}{!}{
		\begin{tabu}{cccccccccc|ccccc|cccc}
			\cline{2-19} 
			
			& \multicolumn{9}{|c|}{MSNO} &\multicolumn{5}{c|}{MSSA}                                                                                                                                                                                   & \multicolumn{4}{c|}{\multirow{2}{*}{DIV2K Val 100 \cite{div2k}}}                                                                                                                                                                      \\ 
			& \multicolumn{5}{|c}{Num of convolutions} & \multicolumn{2}{c}{FEM}& \multicolumn{2}{c|}{SIM}  & \multicolumn{5}{c|}{Num of projection matrices} & \multicolumn{4}{c|}{}\\
			& \multicolumn{1}{|c}{1}                        & \multicolumn{1}{c}{2}                        & \multicolumn{1}{c}{4}                        & \multicolumn{1}{c}{8}                        & 16                 & \multicolumn{1}{c}{W}                         & W/O                      & \multicolumn{1}{c}{W}                         & W/O          & \multicolumn{1}{c}{1 }                        & \multicolumn{1}{c}{2}                        & \multicolumn{1}{c}{4}                         & \multicolumn{1}{c}{8}                        & 16                                      & $\times 2$     & \multicolumn{1}{c}{$\times 4$}               & \multicolumn{1}{c}{$\times 6$}               & \multicolumn{1}{c|}{$\times12$}                                                   \\ \cline{2-19} 
			\specialrule{0em}{1pt}{0pt}
			\cline{2-19}
			\specialrule{0em}{1pt}{0pt}
			
			& \multicolumn{1}{|c}{ \checkmark }                        & \multicolumn{1}{c}{}                         & \multicolumn{1}{c}{}                         & \multicolumn{1}{c}{}                         &              & \multicolumn{1}{|c}{ \checkmark }                         &                          & \multicolumn{1}{|c}{ \checkmark }                         &                           & \multicolumn{1}{c}{}                         & \multicolumn{1}{c}{}                         & \multicolumn{1}{c}{ \checkmark }                         & \multicolumn{1}{c}{}                         &                         &          \multicolumn{1}{c|}{34.83}        & \multicolumn{1}{c|}{29.14}                         & \multicolumn{1}{c|}{26.86}                         & \multicolumn{1}{c|}{23.79}                                                             \\ \cline{2-19} 
			& \multicolumn{1}{|c}{}                         & \multicolumn{1}{c}{ \checkmark }                        & \multicolumn{1}{c}{}                         & \multicolumn{1}{c}{}                         &              & \multicolumn{1}{|c}{ \checkmark }                         &                          & \multicolumn{1}{|c}{ \checkmark }                         &                   & \multicolumn{1}{c}{}                         & \multicolumn{1}{c}{}                         & \multicolumn{1}{c}{ \checkmark }                         & \multicolumn{1}{c}{}                         &                           & \multicolumn{1}{c|}{34.84}                    & \multicolumn{1}{c|}{29.14}                         & \multicolumn{1}{c|}{26.87}                         & \multicolumn{1}{c|}{23.80}                                                          \\ \cline{2-19} 
			& \multicolumn{1}{|c}{}                         & \multicolumn{1}{c}{}                         & \multicolumn{1}{c}{ \checkmark }                        & \multicolumn{1}{c}{}                         &               & \multicolumn{1}{|c}{ \checkmark }                         &                          & \multicolumn{1}{|c}{\checkmark}                         &                  & \multicolumn{1}{c}{}                         & \multicolumn{1}{c}{}                         & \multicolumn{1}{c}{ \checkmark }                         & \multicolumn{1}{c}{}                         &                                    & \multicolumn{1}{c|}{34.84}            & \multicolumn{1}{c|}{29.14}                         & \multicolumn{1}{c|}{26.87}                         & \multicolumn{1}{c|}{23.80}                                                               \\ \cline{2-19} 
			& \multicolumn{1}{|c}{}                         & \multicolumn{1}{c}{}                         & \multicolumn{1}{c}{}                         & \multicolumn{1}{c}{ \checkmark }                        &                & \multicolumn{1}{|c}{ \checkmark }                         &                      & \multicolumn{1}{|c}{ \checkmark}                         &                      & \multicolumn{1}{c}{}                         & \multicolumn{1}{c}{}                         & \multicolumn{1}{c}{ \checkmark }                         & \multicolumn{1}{c}{}                         &                           & \multicolumn{1}{c|}{34.85}               & \multicolumn{1}{c|}{29.15}                         & \multicolumn{1}{c|}{26.88}                         & \multicolumn{1}{c|}{23.80}                                                             \\ \cline{2-19} 
						& \multicolumn{1}{|l}{} & \multicolumn{1}{c}{} & \multicolumn{1}{c}{} & \multicolumn{1}{c}{} & \checkmark & \multicolumn{1}{|c}{\checkmark } &  & \multicolumn{1}{|c}{\checkmark }  &   & \multicolumn{1}{c}{} & \multicolumn{1}{c}{} & \multicolumn{1}{c}{ \checkmark } & \multicolumn{1}{c}{} &  & \multicolumn{1}{l|}{$\textbf{34.87}$} & \multicolumn{1}{l|}{$\textbf{29.16}$} & \multicolumn{1}{l|}{$\textbf{26.89}$} & \multicolumn{1}{l|}{$\textbf{23.81}$}  \\ 
					\cline{2-19}
			\specialrule{0em}{1pt}{0pt}
			\cline{2-19}
			& \multicolumn{1}{|c}{}                         & \multicolumn{1}{c}{}                         & \multicolumn{1}{c}{}                         & \multicolumn{1}{c}{}                         &  \checkmark                  & \multicolumn{1}{|c}{ \checkmark }                         &                          & \multicolumn{1}{|c}{ \checkmark }                         &                 & \multicolumn{1}{c}{ \checkmark }                        & \multicolumn{1}{c}{}                         & \multicolumn{1}{c}{}                          & \multicolumn{1}{c}{}                         &                           & \multicolumn{1}{c|}{$\textbf{34.87}$}                 & \multicolumn{1}{c|}{29.15}                         & \multicolumn{1}{c|}{26.88}                         & \multicolumn{1}{c|}{23.79}                                         \\ \cline{2-19} 
			& \multicolumn{1}{|c}{}                         & \multicolumn{1}{c}{}                         & \multicolumn{1}{c}{}                         & \multicolumn{1}{c}{}                         & \checkmark                          & \multicolumn{1}{|c}{ \checkmark }                         &                          & \multicolumn{1}{|c}{ \checkmark }                         &                  & \multicolumn{1}{c}{}                         & \multicolumn{1}{c}{ \checkmark }                        & \multicolumn{1}{c}{}                          & \multicolumn{1}{c}{}                         &                 & \multicolumn{1}{c|}{$\textbf{34.87}$}                  & \multicolumn{1}{c|}{29.15}                         & \multicolumn{1}{c|}{$\textbf{26.89}$}                         & \multicolumn{1}{c|}{$\textbf{23.81}$}                                                         \\  \cline{2-19} 
			& \multicolumn{1}{|l}{} & \multicolumn{1}{c}{} & \multicolumn{1}{c}{} & \multicolumn{1}{c}{} & \checkmark & \multicolumn{1}{|c}{\checkmark } &  & \multicolumn{1}{|c}{\checkmark }  &   & \multicolumn{1}{c}{} & \multicolumn{1}{c}{} & \multicolumn{1}{c}{ \checkmark } & \multicolumn{1}{c}{} &  & \multicolumn{1}{l|}{$\textbf{34.87}$} & \multicolumn{1}{l|}{$\textbf{29.16}$} & \multicolumn{1}{l|}{$\textbf{26.89}$} & \multicolumn{1}{l|}{$\textbf{23.81}$}  \\ 
			\cline{2-19} 
			& \multicolumn{1}{|c}{}                         & \multicolumn{1}{c}{}                         & \multicolumn{1}{c}{}                         & \multicolumn{1}{c}{}                         &  \checkmark                  & \multicolumn{1}{|c}{ \checkmark }                         &                          & \multicolumn{1}{|c}{ \checkmark }                         &              & \multicolumn{1}{c}{}                         & \multicolumn{1}{c}{}                         & \multicolumn{1}{c}{}                          & \multicolumn{1}{c}{ \checkmark }                        &                               & \multicolumn{1}{c|}{$\textbf{34.87}$}                & \multicolumn{1}{c|}{29.15}                         & \multicolumn{1}{c|}{26.88}                         & \multicolumn{1}{c|}{23.79}                                                     \\ 
			\cline{2-19} 
			& \multicolumn{1}{|c}{}                         & \multicolumn{1}{c}{}                         & \multicolumn{1}{c}{}                         & \multicolumn{1}{c}{}                         &  \checkmark                 & \multicolumn{1}{|c}{ \checkmark }                         &                          & \multicolumn{1}{|c}{ \checkmark }                         &                       & \multicolumn{1}{c}{}                         & \multicolumn{1}{c}{}                         & \multicolumn{1}{c}{}                          & \multicolumn{1}{c}{  }                        &  \checkmark                         & \multicolumn{1}{c|}{34.83}            & \multicolumn{1}{c|}{29.13}                         & \multicolumn{1}{c|}{26.86}                         & \multicolumn{1}{c|}{23.79}                         \\
			\cline{2-19}
			\specialrule{0em}{1pt}{0pt}
			\cline{2-19}
			& \multicolumn{1}{|c}{}                         & \multicolumn{1}{c}{}                         & \multicolumn{1}{c}{}                         & \multicolumn{1}{c}{}                         &  \checkmark              & \multicolumn{1}{|c}{}                          &  \checkmark                         & \multicolumn{1}{|c}{ \checkmark }                         &                 & \multicolumn{1}{c}{}                         & \multicolumn{1}{c}{}                         & \multicolumn{1}{c}{ \checkmark }                         & \multicolumn{1}{c}{}                         &                            & \multicolumn{1}{c|}{$\textbf{34.87}$}               & \multicolumn{1}{c|}{29.15}                         & \multicolumn{1}{c|}{26.88}                         & \multicolumn{1}{c|}{23.80}                                 \\
			
			\cline{2-19} 
			 \multirow{-12}{*}{\rotatebox{90}{EDSR-baseline-MSIT}}                      & \multicolumn{1}{|c}{}             & \multicolumn{1}{l}{}             & \multicolumn{1}{c}{}                         & \multicolumn{1}{c}{}                         &  \checkmark                & \multicolumn{1}{|c}{\checkmark}                          &                           & \multicolumn{1}{|c}{ }                         &        \checkmark              & \multicolumn{1}{c}{}                         & \multicolumn{1}{c}{}                         & \multicolumn{1}{c}{ \checkmark }                         & \multicolumn{1}{c}{}                         &                           & \multicolumn{1}{c|}{34.82}          & \multicolumn{1}{c|}{29.11}                         & \multicolumn{1}{c|}{26.84}                         & \multicolumn{1}{c|}{23.77}                                            \\
			
			\cline{2-19} 

	\end{tabu}}
	
	\caption{Ablation experiments within the modules were conducted on the DIV2K validation set \cite{div2k} for both MSNO and MSSA individually. The best performing results are highlighted in $\textbf{bold}$.}
	\label{tab_comparison_ab3}
\end{table*}
\begin{table}[h]
	\small
	\setlength{\tabcolsep}{1.8mm}
	\centering
	\begin{tabu}{|c|cccc|}
		\hline
		\multicolumn{1}{|c|}{\multirow{2}{*}{Sampling distribution}} & \multicolumn{4}{c|}{DIV2K val 100 \cite{div2k}}                                              \\  
		\multicolumn{1}{|c|}{}                                     & \multicolumn{1}{c}{$\times2$} & \multicolumn{1}{c}{$\times4$} & \multicolumn{1}{c}{$\times6$} & $\times12$ \\ 
		\hline
		\specialrule{0em}{1pt}{0pt} 
		\hline
		Original EDSR-MSIT                & \multicolumn{1}{l|}{{34.87}}          & \multicolumn{1}{l|}{29.16}          & \multicolumn{1}{l|}{26.89}          &    23.81       \\ \hline
		Training with $\eta \thicksim U (1, 4)$                                 & \multicolumn{1}{l|}{$\textbf{34.91}$}          & \multicolumn{1}{l|}{29.19}          & \multicolumn{1}{l|}{26.91}          &    23.83        \\ \hline
		Training with $\eta \thicksim U (1, 8)$                                  & \multicolumn{1}{l|}{34.85}          & \multicolumn{1}{l|}{$\textbf{29.20}$}          & \multicolumn{1}{l|}{$\textbf{26.94}$}          &   $\textbf {23.86}$         \\ \hline
		Training with $\eta \thicksim U (4, 12)$                                & \multicolumn{1}{l|}{34.50}          & \multicolumn{1}{l|}{29.15}          & \multicolumn{1}{l|}{26.93}          &    23.85        \\ \hline
		Training with $\eta \thicksim U (1, 12)$                                & \multicolumn{1}{l|}{34.80}          & \multicolumn{1}{l|}{29.18}          & \multicolumn{1}{l|}{$\textbf{26.94}$}          &   23.85        \\ \hline
	\end{tabu}
	\caption{PSNR (dB) results for the training strategy based on re-parameterization. The best performing results are highlighted in $\textbf{bold}$.}
	\label{tab_comparison_ab2}
\end{table}

\subsection{Ablation studies}
\begin{figure*}[t]
	\centering
	\begin{subfigure}{1\linewidth}
		\centering
		\includegraphics[width=1\linewidth]{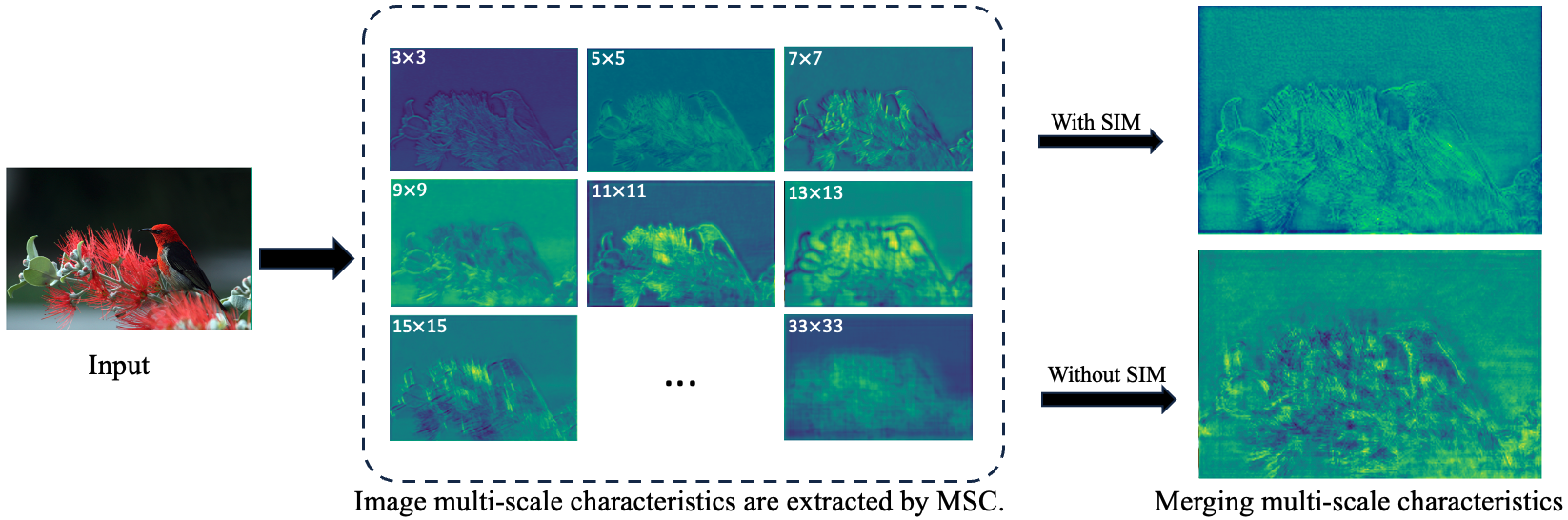}
	\end{subfigure}
	\caption{Visualization experiments for multi-scale characteristics capture and merging.}
	\label{msc_sim}
\end{figure*}
\begin{figure}[h]
	\centering
	\begin{subfigure}{1\linewidth}
		\centering
		\includegraphics[width=1\linewidth]{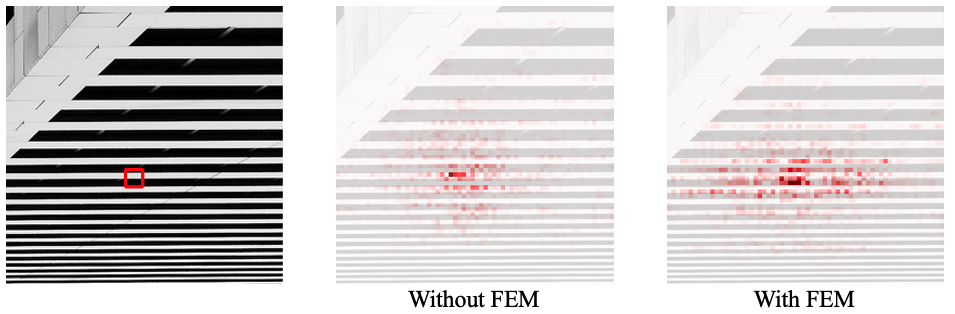}
	\end{subfigure}
	\caption{Visualization experiments using LAM for feature enhancement of FEM.}
	\label{lem}
\end{figure}

In the ablation experiments, we studied each module in MSIT and the submodules within each module. For simplicity, we used the DIV2K dataset with a batch size of 16, otherwise consistent with the above.
\paragraph{Effectiveness of ANSNO}
We employ MSNO to obtain multi-scale characteristics to latent codes in ASSR. To evaluate the effectiveness of MSNO, we conduct a series of comprehensive experiments. As shown in Table \ref{tab_comparison_ab1}, for the fairness of the experiments, we controlled the parameters of the network without MSNO, the addition of MSNO improves the multi-scale characteristics of the latent codes in the network, and therefore the SR performance of each magnifications is improved.
As shown in Fig. \ref{msc_sim}, we visualized the feature maps of the output of each parallel convolutions in MSC. It is worth noting that in the first few convolutions, the detailed texture of the target object is mainly extracted, and with the increase of convolution kernel size, the overall shape of the target object is gradually preserved. Since small SR magnification factors (e.g., $\times$2) are more interested in the overall shape about the target, and large SR magnifications are more inclined to the detailed texture of the image, the performance of the network is improved across all magnification factors.
To further validate the effectiveness of MSNO, we have systematically studied this module by changing the number of parallel convolutions with different convolution kernel sizes in MSC. The results in Tab. \ref{tab_comparison_ab3} indicate that as the number of parallel convolutions increases, so does the performance of the network, and the optimal performance is achieved when the number of parallel convolution is set to 16, corresponding to convolution kernel size of 33$\times$33. Considering our input patch size is 48$\times$48, this kernel size is the maximum. This also reflects the fact that more multi-scale modulation is utilized leads to better performance in ASSR.
We explore other sub-modules within MSNO and systematically demonstrate their individual efficacy. To illustrate the improved feature utilization of FEM, we illustrate this using a Local Attribution Maps (LAM) \cite{lam}. In these maps, areas of interest are represented as red rectangles, and red particles represent the contextual feature used by these areas. The darker the red hue, the greater the utilization of feature. According to Fig. \ref{lem} and Tab. \ref{tab_comparison_ab3}, it can be seen that FEM enriches the information of latent codes without adding additional complexity, thus improving the performance of SR.
In addition we have also evaluated the effect of SIM. According to Tab. \ref{tab_comparison_ab3}, it can be found that the introduction of SIM merges the multi-scale latent codes in the network, and the performance is improved. To further understand this process, we used SIM and regular convolution respectively to merge multi-scale latent codes for visualization experiments, as shown in Fig. \ref{msc_sim}. Regular convolution loses many detail, while SIM avoids this shortcoming by retaining high-frequency features and detailed texture, and fully merging multi-scale latent codes.

\begin{figure*}[h]
	\centering
	\begin{subfigure}{1\linewidth}
		\centering
		\includegraphics[width=1\linewidth]{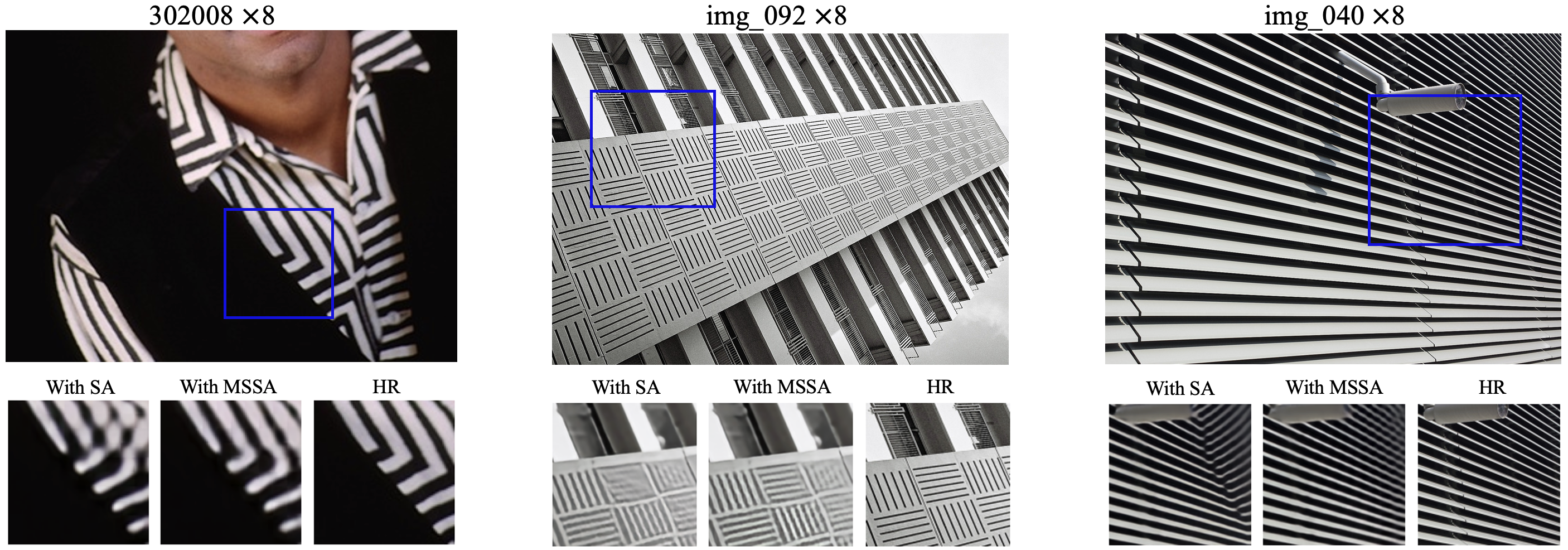}
	\end{subfigure}
	\caption{Qualitative experiments with MSSA vs. conventional SA \cite{clit}}
	\label{mssa_ab}
\end{figure*}
\paragraph{Effectiveness of MSSA}
In order to further focus on multi-scale characteristics, we introduce MSSA. To evaluate the effectiveness of MSSA, we compare it with conventional SA in a quantitative analysis. As depicted in the Tab. \ref{tab_comparison_ab1}, in MSSA column, symbol "$\usym{2613}$" denotes the use of conventional SA. According to the results, using MSSA not only reduces the number of parameters but also improves performance. For $\times4, \times6, \times12$ SR, the network using MSSA outperforms the network using conventional SA. This observation highlights the fact that MSSA is able to more efficiently and comprehensively improve the multi-scale characteristics to handle different magnifications of SR. In addition we further added qualitative experiments to visualize the effect of MSSA, as shown in Fig. \ref{mssa_ab}. In BSD100 302008 \cite{BSD100} and Urban100 img\_092 \cite{U100}, the networks using SA show over-smoothed results, which is alleviated by using MSSA. In Urban100 img\_040, the network using SA shows severe artifacts at the high frequency edges, leading to poor results, while using MSSA avoids this problem by further enhancing multi-scale latent codes.
In addition, we explore the impact of changing the number of projection matrix in MSSA, and the experimental results are presented in the Tab. \ref{tab_comparison_ab3}. It is observed that the best performance is achieved when the number of projection matrix in MSSA is set to 4.

\paragraph{Effectiveness of RIM}
In order to improve performance of the network for different magnification factors of SR, we employ a re-parameterization method based on cumulative training strategy. To validate the effectiveness of RIM, a comparative experiment was conducted, as illustrated in the Tab. \ref{tab_comparison_ab1}. According to the results, it can be found that RIM improves the connections of large convolution weights compared to RefConv \cite{refconv}, and obtains better performance in $\times6, \times12$ SR. To further investigate the use of RIM in cumulative training strategies, we conducted experiments with different training ranges, as shown in Tab. \ref{tab_comparison_ab2}. According to the results, it can be found that the performance can be improved based on the re-parameterization training strategy, such as $\eta \thicksim U(1, 4)$, and the network complexity can be reduced, but for a more balanced network, we choose $\eta \thicksim U(1, 8)$.

	\section{Conclusion}
	In this paper, we propose Multi-Scale Implicit Transformer (MSIT) for arbitrary-scale super-resolution (ASSR). MSIT consists of Multi-Scale Neural Operator (MSNO) and Multi-Scale Self-Attention (MSSA). MSNO consists of Feature Enhancement Module, Multi-Scale Convolution and Scale Integration Module for enriching features, extracting and merging multi-scale characteristics to obtain multi-scale latent codes. MSSA computes attention latent codes by aggregating projection matrices of different scales to further enhance multi-scale latent codes. To improve the performance of the network for different magnification factors, we introduce Re-Interaction Module (RIM) combined with cumulative training strategy. RIM remaps the loaded weights to improve the diversity of learned information by establishing connections among network weights. In contrast to other methods, we systematically introduce multi-scale characteristics into ASSR for the first time, and extensive experimental results demonstrate that MSIT outperforms existing methods and achieves state-of-the-art performance. While MSIT has shown excellent performance, there are some undeniable drawbacks: the model still has a large number of parameters, making it challenging to integrate into embedded devices. In the future, we hope to develop lighter and more performant networks suitable for integration into mobile devices.

	{\small
		\bibliographystyle{unsrt}
		\bibliography{egbib.bib}
	}
	
	\bibliographystyle{unsrtnat}
	
\end{document}